\documentclass[10pt,a4paper]{article}
\usepackage{lrec}
\usepackage{graphicx}
\graphicspath{{./figs/}}
\usepackage[latin1]{inputenc}
\usepackage[hidelinks]{hyperref} 
\usepackage{color}
\usepackage{secdot}   
\sectiondot{subsection}
\usepackage{verbatim}
\usepackage{paralist}

\title{Machine Translation of Low-Resource Spoken Dialects:\\Strategies for Normalizing Swiss German}

\name{Pierre-Edouard Honnet$^{*,1}$\thanks{\hspace{-6mm}$^{*}$ Work conducted while the first and second authors were at the Idiap Research Institute, Martigny, Switzerland.}, 
Andrei Popescu-Belis$^{*,2}$, Claudiu Musat$^3$, Michael Baeriswyl$^3$\\}

\address{
\begin{tabular}{ccc}
$^1$\,Telepathy Labs & $^2$\,HEIG-VD / HES-SO      & $^3$\,Swisscom (Schweiz) AG\\
Sch\"utzengasse 25   & Route de Cheseaux 1, CP 521 & Genfergasse 14\\
CH-8001 Z\"urich     & CH-1401 Yverdon-les-Bains   & CH-3011 Bern \\
Switzerland          & Switzerland                 & Switzerland \\
\texttt{\small{pierre-edouard.honnet}} & \texttt{\small{andrei.popescu-belis}} & \texttt{\small{claudiu.musat@swisscom.com}}\\
\texttt{\small{@telepathy.ai}}
                               & \texttt{\small{@heig-vd.ch}}& \texttt{\small{michael.baeriswyl@swisscom.com}}\\
\end{tabular}\\}
			
\abstract{The goal of this work is to design a machine translation (MT) system for a low-resource family of dialects, collectively known as Swiss German, which are widely spoken in Switzerland but seldom written.  We collected a significant number of parallel written resources to start with, up to a total of about 60k words.  Moreover, we identified several other promising data sources for Swiss German.  Then, we designed and compared three strategies for normalizing Swiss German input in order to address the regional diversity.  We found that character-based neural MT was the best solution for text normalization.  In combination with phrase-based statistical MT, our solution reached 36\% BLEU score when translating from the Bernese dialect.  This value, however, decreases as the testing data becomes more remote from the training one, geographically and topically.  These resources and normalization techniques are a first step towards full MT of Swiss German dialects.\\ \newline 
\Keywords{machine translation, low-resource languages, spoken dialects, Swiss German, character-based neural MT} }

\begin{document}

\maketitleabstract

\section{Introduction}

In the era of social media, more and more people make online contributions in their own language.  The diversity of these languages is however a barrier to information access or aggregation across languages. Machine translation (MT) can now overcome this limitation with considerable success for well-resourced languages, i.e.\ language pairs which are endowed with large enough parallel corpora to enable the training of neural or statistical MT systems.  
This is not the case, though, for many low-resourced languages which have been traditionally considered as oral rather than written means of communication, and which often lack standardized spelling and/or exhibit significant variations across dialects.  
Such languages have an increasing presence in written communication, especially through social media, while remaining inaccessible to non-speakers.

This paper presents a written MT system for a mostly spoken family of dialects: Swiss German.  Although spoken in a technologically developed country by around five million native speakers, Swiss German has never been significantly used in writing -- with the exception of folklore or children books -- 
before the advent of social media.  Rather, from primary school, speakers of Swiss German are taught to use High German in writing, more precisely a variety known to linguists as Swiss Standard German, which is one of the three official federal languages along with French and Italian.  Swiss German is widely used in social media, but foreigners or even Swiss speakers of the other official languages cannot understand it.

In this paper, we describe the first end-to-end MT system from Swiss German to High German.  In Section~\ref{sec:gsw-resources}, we present the Swiss German dialects and review the scarce monolingual and even scarcer parallel language resources that can be used for training MT. In Section~\ref{sec:previous-work}, we review previous work on Swiss German and on MT of low-resource languages. In Section~\ref{sec:normalization-gsw}, we address the major issue of dialectal variation and lack of standard spelling -- which affects many other regional and/or spoken languages as well -- through three solutions: explicit conversion rules, phonetic representations, and character-based neural MT.  These solutions are combined with phrase-based statistical MT to provide a standalone translation system, as explained in Section~\ref{sec:gsw-mt}.  In Section~\ref{sec:results} we present evaluation results. We first find that the similarity between the regions of training vs.\ test data has a stronger effect on performance than the similarity of text genre.  Moreover, the results show that character-based NMT is beneficial for dealing with spelling variation.  Our system is thus an initial general purpose MT system making Swiss German accessible to non-speakers, and can serve as a benchmark for future, better-resourced attempts.


\section{Collecting Swiss German Resources}
\label{sec:gsw-resources}

\subsection{A Heterogeneous Family of Dialects}

\paragraph{Definition.} Swiss German \cite{russ1990dialects,christen2013sprachatlas} is a family of dialects used mainly for spoken communication by about two thirds of the population of Switzerland (i.e.\ over five million speakers).  Swiss German is typically learned at home as a first language, but is substituted starting from primary school by High German for all written forms, as well as for official spoken discourse, for instance in politics or the media.  Linguistically, the variety of High German written and spoken in Switzerland is referred to as \emph{Swiss Standard German} (see Russ \shortcite{russ1994german}, Chapter~4, p.\ 76--99) and is almost entirely intelligible to German or Austrian speakers.  On the contrary, Swiss German is generally not intelligible outside Switzerland.

In fact, Swiss German constitutes a group of heterogeneous dialects, which exhibit strong local variations.  Due to their spoken nature, they have no standardized written form: for instance, the word \emph{kleine} (meaning  small' in Standard German) could be written as \emph{chlyni}, \emph{chliini}, \emph{chline}, \emph{chli} or \emph{chlii} in Swiss German.  Linguistic studies of the Swiss German dialects (see Russ \shortcite{russ1990dialects} or Christen et al. \shortcite{christen2013sprachatlas}) generally focus on the phonetic, lexical or syntactic variations and their geographical distribution, often concluding that such variations are continuous and non-correlated with each other.  Finally, little teaching material in Swiss German is available to foreigners.

\begin{figure}[t]
\centering
\includegraphics[width=0.85\linewidth]{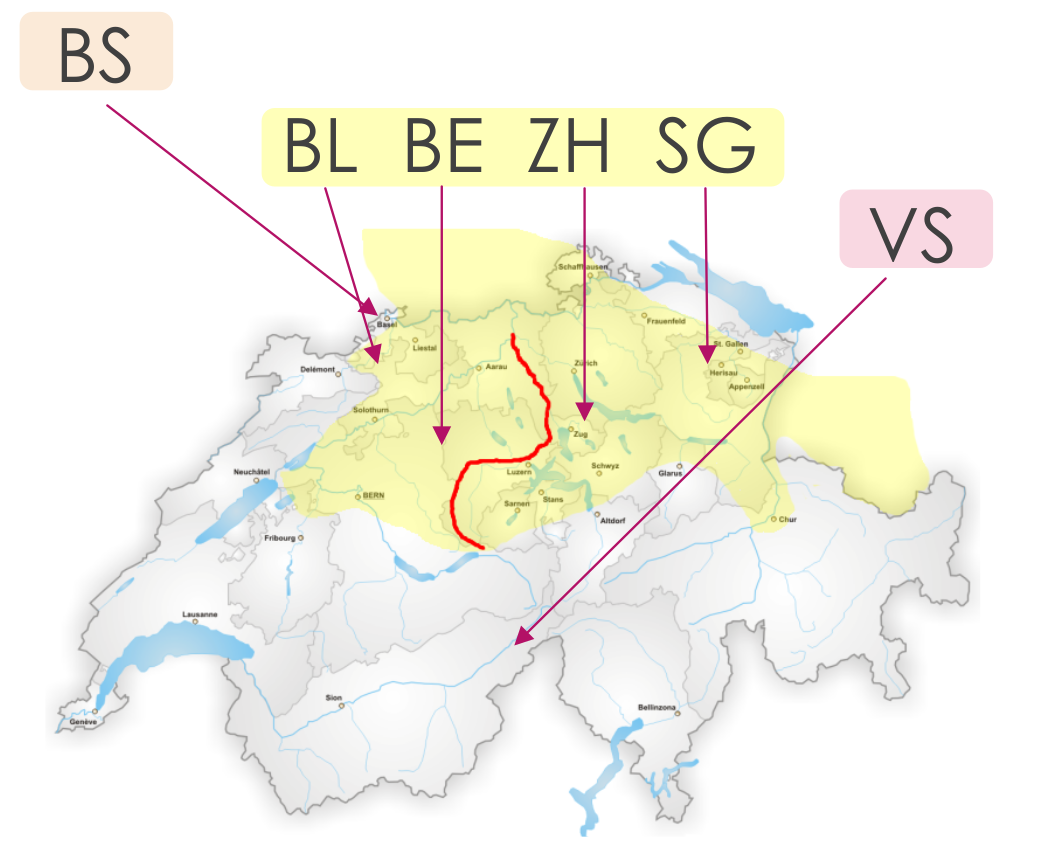}
\caption{Map of Switzerland, with six main dialects
  that we identify for our study. The area in yellow
  indicates the High Alemannic dialects. Image source: https://commons.wikimedia.org/wiki/File:Brunig-Napf-Reuss-Linie.png.}
\label{fig:swiss_map}
\end{figure}

\paragraph{Divisions.} The areas where each dialect is spoken are influenced both by the administrative divisions (cantons and communes) and by natural borders (topography).  Within the large group of Germanic languages, the dialects of Switzerland belong to the \emph{Alemannic} group.  However, while a majority of dialects are \emph{High Alemannic} (yellow area on map in Figure~\ref{fig:swiss_map}), those spoken in the city of Basel and in the Canton of Valais belong respectively to the \emph{Low Alemannic} and the \emph{Highest Alemannic} groups.  Within the High Alemannic group, a multitude of divisions have been proposed.  One of the most consistent ones is the Brunig-Napf-Reuss line between the eastern and western groups (red line in Fig.~\ref{fig:swiss_map}).  A fine-grained approach could easily identify one or more dialects for each canton.

For the purpose of this study, we distinguish only two additional sub-groups on each side of the Brunig-Napf-Reuss line, and refer to them using the largest canton in which they are spoken.  Westwards, we distinguish the Bernese group from the group spoken around Basel (cantons of Basel-Country, Solothurn and parts of Aargau).  Eastwards, we distinguish the Z\"{u}rich group from the easternmost group around St.~Gallen.  Therefore, for training and testing MT on various dialects, we consider in what follows six main variants of Swiss German, represented on the map in Figure~\ref{fig:swiss_map}.  

\paragraph{Notations.} We refer to Swiss German as `GSW' (abbreviation from ISO 639-2) followed by the indication of the variant:  GSW-BS (city of Basel), GSW-BL (regions of Basel, Solothurn, parts of Aargau), GSW-BE (mainly canton of Bern), GSW-ZH (canton of Zurich and neighbors), GSW-SG (St.~Gallen and easternmost part of Switzerland), GSW-VS (the German-speaking part of the canton of Valais/Wallis).  
These groups correspond to the dialect labels used in the Alemannic Wikipedia (see Section~\ref{sec:parallel-resources} below): \emph{Basel}, \emph{Baselbieter}, \emph{Bern}, \emph{Zurich}, \emph{\`{U}ndert\`{o}ggeborg}, and \emph{Wallis} (Valais).
In contrast, Swiss Standard German is referred to as `DE-CH', a qualified abbreviation from IETF.  Moreover, below, we will append the genre of the training data to the dialect abbreviation.

\paragraph{Usage and Need for MT.} Swiss German is primarily used for spoken communication, but the widespread adoption of social media in Switzerland has significantly increased its written use for informal exchanges on social platforms or in text messages.  No standardized spelling has emerged yet, a fact related to the lack of GSW teaching as a second language.  GSW is still written partly with reference to High German and partly using a phonetic transcription, also inspired from German pronunciation.  Access to such content in social media is nearly impossible to foreigners, and even to speakers of different dialects, e.g.\ Valaisan content to Bernese speakers.  Our goal is to design an MT system translating all varieties of GSW (with their currently observed spelling) towards High German, taking advantage of the relative similarity of these languages.  By pivoting through High German, other target languages can then be supported.  Moreover, if a speech-to-text system existed for Swiss German \cite{Garner2014b}, our system would also enable spoken translation.


\subsection{Parallel Resources}
\label{sec:parallel-resources}

Despite attempts to use comparable corpora or even monolingual data only (reviewed in Section~\ref{sec:previous-work}), parallel corpora aligned at the sentence level are essential resources for training statistical MT systems.  
In our case, while written resources in Swiss German are to some extent available (as reviewed in Section~\ref{sec:monolingual-resources}), it is rare to find their translations into High German or vice-versa.  When these are available, the two documents are often not available in electronic version, which requires a time-consuming digitization effort to make them usable for MT.\footnote{Many of them are children books, such as \emph{Pitschi} by Hans Fischer, \emph{The Gruffalo} by Julia Donaldson, or \emph{The Little Prince} by Antoine de Saint-Exup\'{e}ry.  Other examples include transcripts of Mani Matter's songs, or several Asterix comics in Bernese.}

One of our goals is to collect the largest possible set of parallel GSW/DE texts, in a first stage regardless of their licensing status.  We include among such resources parallel lexicons (``dictionaries''), and show that they are helpful for training MT.  We summarize in Table~\ref{tab:data} the results of our data collection effort, providing brief descriptions of each resource with especially their variant of GSW and their domain.  We describe in detail each resource hereafter. 

\begin{table}[ht]
  \centering
  \begin{tabular}{|l|@{\,}r|r|r|@{\,}r|}
	  \hline
    Dataset & Train & Dev. & Test & Total\\
    \hline
    GSW-BE-Novel     & 2,667  & 218   & 183   & 3,251  \\
    GSW-BE-Wikipedia & --     & 180   & 67    & 247    \\
    GSW-VS-Radio     & 463    & 100   & 50    & 613    \\
    GSW-ZH-Wikipedia & --     & 45    & 50    & 95     \\
    GSW-BE-Bible     & --     & --    & 126   & 126    \\
		GSW-Archimob     & 40,159 & 2,710 & 2,710 & 45,579 \\
		GSW-ZH-Lexicon1  & 1,527  & --    & --    & 1,527  \\ 
		GSW-BE-Lexicon2  & 1,224  & --    & --    & 1,224  \\
		\hline
  \end{tabular}
	\caption{\label{tab:data} GSW/DE parallel datasets partitioned for MT training, tuning and testing, with sizes in numbers of parallel sentences.  The lexicons (last two lines) were not used for testing, and 183 additional lines from GSW\_BE\_Novel are kept apart for future testing.}
\end{table}

\begin{description}

\item[GSW-BE-Novel.] Translations of books from DE into GSW are non-existent. We thus searched for books written originally in GSW and then translated into DE. Among the growing body of literature published in Swiss German, we found only one volume translated into High German and available in electronic form: \emph{Der Goalie bin ig} (in English: \emph{I am the Keeper}), written in Bernese by Pedro Lenz in 2010.  The DE translation stays close to the original GSW-BE text, therefore sentence-level alignment was straightforward, resulting in 3,251 pairs of sentences with 37,240 words in GSW-BE and 37,725 words in DE.


\item[GSW-BE-Wikipedia] and \textbf{GSW-ZH-Wikipedia}. The Ale\-mann\-ic version of Wikipedia\footnote{\url{http://als.wikipedia.org}} appeared initially as a promising source of data.  However, its articles are written not only in Swiss German, but also in other Alemannic dialects such as Alsatian, Badisch and Swabian.  As its contributors are encouraged to write in their own dialects, only a few articles are homogeneous and have an explicit indication of their dialect, using an Infobox with one of the six labels indicated above.  Among them, even fewer have an explicit statement indicating that they have been translated from High German (which would make the useful as parallel texts).  We identified two such pages and sentence-aligned them to serve as test data: ``\emph{Hans Martin Sutermeister}'' translated from DE into GSW-BE and ``\emph{W\"{a}denswil}'' from DE into GSW-ZH.\footnote{These pages are respectively available at
\url{https://de.wikipedia.org/wiki/Hans_Martin_Sutermeister} (High German),
\url{https://als.wikipedia.org/wiki/Hans_Martin_Sutermeister} (Bernese),
\url{https://de.wikipedia.org/wiki/W\%E4denswil} (High German), and
\url{https://als.wikipedia.org/wiki/W\%E4denswil} (Zurich Swiss German).}

\item[GSW-VS-Radio.] A small corpus of Valaisan Swiss German (also called \emph{Wallisertiitsch}) has been collected at the Idiap Research Institute \cite{Garner2014b}.\footnote{\url{www.idiap.ch/dataset/walliserdeutsch}}
The corpus consists of transcriptions of a local radio broadcast\footnote{Radio Rottu, \url{http://www.rro.ch}.} translated into High German.

\item[GSW-BE-Bible.] The Bible has been translated in several GSW dialects, but the only electronic version available to us were online excerpts in Bernese.\footnote{\url{www.edimuster.ch/baernduetsch/bibel.htm}}  However, this is not translated from High German but from a Greek text, hence the alignment with any of the German Bibles is problematic.\footnote{ \url{www.die-bibel.de/bibeln/online-bibeln/}}  We selected the contemporary \emph{Gute Nachricht Bibel} (1997) for its modern vocabulary, and generated parallel data from four excerpts of the Old and New Testament, while acknowledging their particular style and vocabulary.  The following excerpts were aligned: \textit{\"Use Vatter}, \textit{D Wienachtsgschicht}, \textit{Der barmh\"arzig Samaritaner} and \textit{D W\"alt wird erschaffe}.
	
\item[GSW-Archimob.] Archimob is a corpus of standardized Swiss German \cite{Samardzic-LREC-2016}, consisting of transcriptions of interviewees speaking Swiss German, with a word-align normalized version in High German.\footnote{\url{http://www.spur.uzh.ch/en/departments/korpuslab/ArchiMob.html}}  The interviews record memories of WW\,II, and all areas of Switzerland are represented.  In most cases, the normalization provides the corresponding High German word or group of words, but in other cases it is Swiss German with a standardized orthography devised by the annotators.  Using a vocabulary of High German, we filtered out all sentences whose normalizations included words outside this vocabulary.  In other words, we kept only truly High German sentences, along with their original Swiss German counterparts, resulting in about 45,000 GSW/DE word-aligned sentence pairs.

\item[GSW-ZH-Lexicon] and \textbf{GSW-BE-Lexicon}.  The last two parallel resources are vocabularies, i.e.\ lists of GSW words with their DE translation.  As such, they are useful for training our research systems, but not for testing them.  The first one is based on \emph{Hoi Z\"ame}, a manual of Z\"urich Swiss German intended for High German speakers.  The data was obtained by scanning the printed version, performing OCR\footnote{Tesseract: \url{https://github.com/tesseract-ocr/}} and manually aligning the result.  Although the book contains also parallel sentences, only the bilingual dictionary was used in our study, resulting in 1,527 words with their translations. A similar dictionary for Bernese (GSW-BE vs.\ DE) was found online\footnote{\url{www.edimuster.ch/baernduetsch/woerterbuechli.htm}} with 1,224 words for which we checked and corrected the alignments.
\end{description}

\subsection{Monolingual Resources}
\label{sec:monolingual-resources}

The Phonolex dictionary, a phonetic dictionary of High German,\footnote{\url{www.bas.uni-muenchen.de/forschung/Bas/BasPHONOLEXeng.html}. 
We also use it to find OOV words.} was used for training our grapheme-to-phoneme converter (see Section~\ref{sec:norm-phonetic}). 
It contains High German words with their phonetic transcriptions.  

About 75 pages from the Alemannic Wikipedia mentioned above have been collected and used to derive orthographic normalization rules in Section~\ref{sec:norm-rules}.
To build language models (see Section~\ref{sec:gsw-mt}) we used the News Crawls 2007--2015 from the Workshop on MT.\footnote{\url{http://www.statmt.org/wmt17/translation-task.html}}


\section{Previous Work on Swiss German and the MT of Low-Resource Languages}
\label{sec:previous-work}

The variability of Swiss German dialects has been investigated in a number of studies, such as those by Russ \shortcite{russ1990dialects}, Scherer \shortcite{scherrer-thesis-2012}, and Christen et al.\ \shortcite{christen2013sprachatlas}.  This variability was illustrated in a system for generating Swiss German text, with fine-grained parameters for each region on a map \cite{scherrer2012}.

Language resources for Swiss German are extremely rare.  The ArchiMob corpus \cite{Samardzic-LREC-2016} is quite unique, as it provides transcripts of spoken GSW narratives, along with their normalization, as presented above \cite{Samardzic-LTC-2015}.  First performed manually -- thus generating ground-truth data -- the normalization was then performed automatically using character-based statistical MT \cite{Scherrer-KONVENS-2016}.

Initial attempts for MT of GSW include the above-mentioned system for generating GSW texts from DE \cite{scherrer-thesis-2012}, and a system combining ASR and MT of Swiss German from Valais \cite{Garner2014b}.   A normalization attempt for MT, on a different Germanic dialect, has been proposed for Viennese \cite{hildenbrandt2013orthographic}. 

The MT of low-resource languages or dialects has been studied on many other important cases, in particular for Arabic dialects which are also predominantly used for spoken communication \cite{Zbib:2012:MTA}.  The lack of a normalized spelling of dialects has for instance an impact on training and evaluation of automatic speech recognition: a solution is to address spelling variation by mining text from social networks \cite{ali2017werd}.  Other strategies are the crowdsourcing of additional parallel data, or the use of large monolingual and comparable corpora to perform bilingual lexicon induction before training an MT system \cite{klementiev-EtAl:2012:EACL2012,irvine-callisonburch:2013:WMT,irvine2016end}. The METIS-II EU project replaced the need for parallel corpora by using linguistic pre-processing and statistics from target-language corpora only \cite{carl2008metis}.  In a recent study applied to Afrikaans-to-Dutch translation, the authors use a character-based ``cipher model'' and a word-based language model to design a decoder for the low-resourced input language \cite{pourdamghani2017deciphering}.

The Workshops on Statistical MT have proposed translation tasks for low-resourced languages to/from English, such as Hindi in 2014 \cite{bojar-EtAl:2014:W14-33}, Finnish in 2015, or Latvian in 2017.  However, these languages are clearly not as low-resourced as Swiss German, and possess at least a normalized version with a unified spelling.  In 2011, the featured translation task aimed at translating text messages from Haitian Creole into English, with a parallel corpus of similar size as ours (ca.\ 35k words on each side, plus a Bible translation).  The original system built in the wake of the 2010 Haiti earthquake leveraged a phonetic mapping from French to Haitian Creole to obtain a large bilingual lexicon \cite{lewis2010haitian,lewis2011crisis}.


\section{Normalizing Swiss German for MT}
\label{sec:normalization-gsw}

Three issues must be addressed when translating Swiss German into High German, which all contribute to a large number of out-of-vocabulary (OOV) words (i.e.\ previously unseen during training) in the source language:
\begin{compactenum}
\item The scarcity of parallel GSW/DE data for training (see Section~\ref{sec:parallel-resources}), which cannot be easily addressed by the strategies seen in Section~\ref{sec:previous-work}.
\item The variability of dialects across training and testing data, which increases dialect-specific scarcity.
\item The lack of a standard spelling, which introduces intra-dialect and intra-speaker variability. 
\end{compactenum}

There are several ways to address these issues.  The most principled one is the normalization of all GSW input using unified spelling conventions, coupled with the design of a GSW/DE MT system for normalized input.  However, such a goal is far too ambitious for our scope.  Instead, we propose here to normalize Swiss German input for the concrete perspective of MT by converting unknown GSW words either to known GSW ones or to High German ones, which are preserved by the GSW/DE MT system and increase the number of correctly translated words.\footnote{The MT system is specifically built so that OOV words are copied in the target sentence, rather than deleted.}   

This procedure, summarized below, rests on the assumption that many OOV GSW words are close to DE words, but with a slightly different pronunciation and spelling (see examples in the third column of Table~\ref{tab:ortho_rules}).  Each of the three strategies follow the same procedure:
\begin{compactenum}
\item For each OOV word $w$, apply the normalization strategy.  If it changes $w$ into $w'$ then go to (2), if not to (4).
\item If $w'$ is a known GSW word then replace $w$ with $w'$ and proceed to (4), if not, go to (3).
\item If $w'$ is a known DE word then replace $w$ with $w'$.  If not, leave $w$ unchanged and go to (4).
\item Translate the resulting GSW text.
\end{compactenum}
This normalization method has two possible chances to help MT, by converting OOV words either into a known GSW word, or into a correct DE word which is no longer processed by MT.   We describe below three strategies to normalize GSW text input before GSW/DE MT.


\subsection{Explicit Spelling Conversion Rules}
\label{sec:norm-rules}

The first strategy is based on explicit conversion rules for every OOV word $w$, which is changed into $w'$ by applying in sequence several spelling conversion rules, keeping the result if it is a GSW or a DE word, as explained above.  The orthographic rules implemented in our system are shown in Table~\ref{tab:ortho_rules}, with possible conversion examples.

\begin{table}[ht]
  \centering
  \begin{tabular}{|c|c|l|}
	\hline
    Spelling &  Convert to & Example \\
    \hline
    .*scht.*     & .*st.*      & Angscht $\rightarrow$ Angst \\
    .*schp.*     & .*sp.*      & Schprache $\rightarrow$  Sprache \\
    \^{}g\"age.* & \^{}gegen.* & G\"agesatz $\rightarrow$ Gegensatz \\
    C\"aC        & CeC         & Pr\"asident $\rightarrow$ President \\
    \^{}gm.*     & \^{}gem     & Gmeinde $\rightarrow$ Gemeinde \\
    \^{}gf.*     & \^{}gef     & gfunde $\rightarrow$ gefunde(n) \\
    \^{}gw.*     & \^{}gew     & gw\"ahlt $\rightarrow$ gew\"ahlt \\
    \^{}aa.*     & \^{}an.*    & Aafang $\rightarrow$ Anfang \\
    .*ig\$       & .*ung\$     & Regierig $\rightarrow$ Regierung \\
    \^{}ii.*     & \^{}ein.*   & Iiwohner $\rightarrow$ Einwohner \\
		\hline
  \end{tabular}
	\caption{\label{tab:ortho_rules} Orthographic conversion rules using  meta-characters \^{} and \$ for the beginning and end of a word, .* for any sequence of characters, and C for any consonant.} 
\end{table}


\subsection{Using Phonetic Representations}
\label{sec:norm-phonetic}

The second approach is based on the assumption that despite
spelling differences, variants of the same word will have the same pronunciation.  Thus, converting an out-of-vocabulary (OOV) word to its phonetic transcription may allow finding the equivalent word which is present in the vocabulary.  In this case, substituting the OOV word with a known word with the same pronunciation should help MT, assuming the same meaning.

For this, a grapheme-to-phoneme (G2P) converter is needed.  It
consists of an algorithm which is able to convert sequences of characters
into phonetic sequences, or go from the written form of a word to its
pronunciation.  The idea is to build it on High German, as we
expect Swiss German to be written in a phonetic way, which means that
the G2P conversion should be close to High German
pronunciation rules.  In our experiments, a G2P converter was trained
on the Phonolex dictionary, which contains High German words with their 
phonetic transcriptions.  A GSW phonetic dictionary was created by
using this system.  To translate a new OOV word, we convert the word to its pronunciation, and check whether the resulting pronunciation exists either in the phonetic GSW dictionary or in the phonetic DE dictionary, following the procedure explained at the beginning of this section.


\subsection{Character-based Neural MT}
\label{sec:norm-cbnmt}

Mainstream neural MT systems are typically trained using recurrent neural networks (RNNs) to encode a source sentence, and then decode its representation into the target language \cite{cho2014learning}.  The RNNs are often augmented with an attention mechanism to the source sentence \cite{bahdanau2014neural}.  However, training an NMT is not feasible for GSW/DE, as the size of our resources is several orders of magnitude below NMT requirements.  However, several recent approaches have explored a new strategy: the translation system is trained at the character level \cite{ling2015character,costa2016character,chung2016character,bradbury2016quasi,lee2017fully}, or at least character-level techniques such as byte-pair encoding are used to translate OOV words \cite{sennrich2016neural}.

As the available data is limited, one possible approach is to combine a PBSMT and a CBNMT system: the former translates known words,  while the latter translates OOV ones.  The CBNMT system has two main advantages: it can translate unseen words based on the spelling regularities observed in the training data, and it can be trained with smaller amounts of data compared to the requirements of standard NMT methods.

Among the CBNMT approaches, we use here Quasi Recurrent Neural Networks (QRNNs) \cite{bradbury2016quasi}, which take advantage of both convolutional and recurrent layers.  The increased parallelism introduced by the use of convolutional layers allows to speed up both training and testing of translation models.  There are two advantages to use CBNMT for OOV translation only.  First, the training data may be sufficient to capture spelling conversion better than hand-crafted rules such as those in Table~\ref{tab:ortho_rules}.  Second, we can use smaller recurrent layers, as the character sequences to translate for OOV words are much shorter than sentences.

We built a CBNMT system for OOV words based on open source scripts for TensorFlow available online, using the implementation of the QRNN architecture proposed by Kyubyong Park,\footnote{\url{https://github.com/Kyubyong/quasi-rnn}}
with the following modifications:
\begin{compactenum}
\item We added a ``start of word'' symbol to avoid mistakes on the first letter of the word. This was done outside the translation
  scripts, by adding the `:' symbol to each word before the first
  letter, and removing it after translation.
\item We modified the QRNN translation script to allow the translation of input texts without scoring the translation (for the production mode, when no reference is available).
\item We added the possibility to translate an incomplete minibatch, by padding the last incomplete batch with empty symbols~(0).\footnote{Originally, if the size of the minibatch is $n$, and the number of sentences modulo $n$ is $y$ (i.e.\ there are $n*x + y$ sentences), then only the $n*x$ first sentences were translated by the system, which ignored the $y$ last ones.}
\item We set the following hyper-parameters: the maximum number of characters is 40, as no longer words were found in our GSW vocabulary.  The minibatch size was kept to 16, and the number of hidden units was kept to 320, as in the default implementation.
\end{compactenum}
We trained the CBNMT model using unique word pairs from the Archimob corpus (see~\ref{sec:parallel-resources} above), i.e.\ a Swiss German word and its normalized version, with a training set of 40,789 word pairs and a development set of 2,780 word pairs.


\section{Integration with Machine Translation}
\label{sec:gsw-mt}

We use phrase-based statistical MT for the core of our system, as data was not sufficient to train neural MT.  We experimented indeed with two NMT systems\footnote{The DL4MT toolkit \url{https://github.com/nyu-dl/dl4mt-tutorial} and the OpenNMT-py one \url{https://github.com/OpenNMT/OpenNMT-py}.}, which are typically trained on at least one million sentences, and tuned on 100k.  In our case, the available data did not allow NMT to outperform PBSMT, which is used below.

Using the Moses toolkit\footnote{\url{http://www.statmt.org/moses/}} to build a PBSMT system \cite{koehn2003statistical}, we used various subsets of the parallel GSW/DE data presented in Section~\ref{sec:parallel-resources} above to learn translation models.  As for the target language model, we trained a tri-gram model using IRSTLM\footnote{\url{http://hlt-mt.fbk.eu/technologies/irstlm}} over ca.\ 1.3 billion words in High German, and tuned the system using the development data indicated above in Section~\ref{sec:monolingual-resources}.  As explained above, the normalization strategies are used to attempt to change GSW OOV words into either GSW or even DE words that are in the vocabulary.  As we will see, we have combined two strategies in several experiments.  We will use the most common metric for automatic MT evaluation, i.e.\ the BLEU score \cite{papineni2002bleu}.


\section{Results and Discussion}
\label{sec:results}


\subsection{Effects of Genre and Dialect}

The first system to translate from Swiss German into High German was built using Moses trained on the Bernese novel corpus (GSW-BE-Novel in Table~\ref{tab:data} above), with character-based NMT for OOV words.
Table~\ref{tab:bleu_bd_baseline} shows the BLEU scores obtained when
testing this system on test sets from different regions or topics. 
Moreover, we also vary the tuning sets, including ones closer to the target test domains to assess their impact.  The best BLEU score is around 35\% which can be compared, for instance, with Google's NMT scores of 41\% for EN/FR and 26\% for EN/DE, trained on tens of millions of sentences and nearly one hundred processors \cite{wu2016google}.  In our case, our modest resources enable us to reach quite a high score thanks to the normalization strategy and the relative similarity of GSW to DE.

\begin{table}[ht]
  \centering
  \begin{tabular}{|l|l|r|}
	\hline
  Test set         & Tuning (dev) set & BLEU  \\
  \hline
  GSW-BE-Novel     & GSW-BE-Novel     & 35.3 \\
  GSW-BE-Wikipedia & GSW-BE-Novel     & 21.9 \\
  ~~~~\emph{same}  & GSW-BE-Wikipedia & 21.7 \\
  GSW-ZH-Wikipedia & GSW-BE-Novel     & 16.2 \\
	~~~~\emph{same}  & GSW-ZH-Wikipedia & 15.3 \\
  GSW-VS-Radio     & GSW-BE-Novel     & 9.7 \\
	\hline
  \end{tabular}
	\caption{\label{tab:bleu_bd_baseline} BLEU scores for various tuning and test sets for the baseline system trained on GSW-BE-Novel.  Performance decreases significantly as the dialect and domain are more remote from the training/tuning data.}
\end{table}

A typical output of this system is:
\begin{compactdesc}
\item[GSW-BE source:] \emph{fasch wiwenernol\"{a}nger h\"{a}tt w\"{o}uek\"{u}sse . oder hanis\"{a}chtnumegmeint?}
\item[DE MT:] \emph{fast wie wenn er noch l\"{a}nger h\"{a}tte wollten k\"{u}sse . oder hab ich es wohl nur gemeint ?}
\item[Human DE reference:] \emph{als h\"{a}tte er noch l\"{a}nger k\"{u}ssen wollen . oder etwa nicht ?}
\end{compactdesc}

The scores in Table~\ref{tab:bleu_bd_baseline} show the following trends:
\begin{compactenum}
\item When testing on similar data, i.e.\ the same dialect and same
  domain, the scores are the highest,
  and in the same range as state of the art EN-DE or EN-FR systems.
\item When changing domain (testing on Wikipedia data in the same
  dialect), the scores are decreasing.
\item When testing on different dialects, the scores decrease
  more. This is true both for GSW-ZH and GSW-VS. As the dialect and
  domain are further from the data used to train the system, the score
  gets lower.  GSW-VS is known to be very different from any other
  GSW dialect, and radio broadcast data is expected to be very different
  from the novel used at training time.
\end{compactenum}

\subsection{Effect of the Size of Training Data and Language Model}
\label{sec:increase_data}

To evaluate first the effect of using more training data, with larger vocabularies, a new system was trained using the same data as in the
previous experiments, complemented with the two bilingual lexicons presented in Section~\ref{sec:parallel-resources}.   Table~\ref{tab:bleu_bd_mixed}, second column, presents the resulting BLEU scores, which increase in all cases by about 1 BLEU point.  As expected, using more training data in the form of bilingual lexicons yields more reliable translation models.

To build more robust systems, we also used a larger target language model built on the NewsCrawl 2007-2015 data from WMT (see Section~\ref{sec:monolingual-resources}) instead of only the DE side of our parallel data, which is still used for training the translation models, as above. 
Table~\ref{tab:bleu_bd_mixed}, third column, gives the BLEU scores on the same test sets, using the larger target language model.  We observe that the scores decrease slightly for the Bernese test sets, and hypothesize that this is due to the different domains of the language model and the test set.  However, as the larger language model is trained on more diverse data, we will keep using it below for its robustness.

\begin{table}[ht]
  \centering
  \begin{tabular}{|l|r|r|}
	\hline
	& \multicolumn{2}{c|}{BLEU}\\
    Test set &  Small LM & Large LM \\
    \hline
    GSW-BE-Novel     & 36.2 & 34.1 \\
    GSW-BE-Wikipedia & 23.6 & 22.7 \\
    GSW-ZH-Wikipedia & 17.3 & 17.7 \\
    GSW-VS-Radio     & 10.0 & 10.7 \\
		\hline
  \end{tabular}
	  \caption{\label{tab:bleu_bd_mixed} BLEU scores for various test sets (Bern, Zurich, Valais dialects) for a Moses-based system trained over data \emph{including the two GSW/DE dictionaries} with two language models (LM).}
\end{table}

\subsection{Out-of-Vocabulary GSW Words}

The three approaches proposed for normalization were evaluated on the same datasets as the previous systems.  Additionally, two other approaches combining, on one side orthographic and phonetic based conversions, and on the other side CBNMT and phonetic conversion, were evaluated.
Table~\ref{tab:bleu_proposed} summarizes the results for the baseline
system and the proposed approaches.  

\begin{table*}[t]
  \centering
  \begin{tabular}{|l|c|c|c|c|c|c|c|}
	\hline
  Test set &  Baseline1 & Baseline2 & Phon.\ & Orth.\ & Orth.\ \& Phon.\ & CBNMT \& Phon.\ & CBNMT\\
  \hline
  GSW-Archimob     & 10.9 & 10.8 & 10.8 & 11.2 & 13.9 & 27.9 &{\bf 32.9} \\
  GSW-BE-Novel     & {\bf 36.2} & 34.1 & 34.3 & 34.4 & 34.4 & 35.6 & 35.4 \\
  GSW-BE-Wikipedia & 23.6 & 22.7 & 23.2 & 23.6 & 23.7 & 20.5 & {\bf 24.0} \\
  GSW-ZH-Wikipedia & 17.3 & 17.7 & 17.1 & 18.9 & 18.2 & 22.0 & {\bf 22.1} \\
  GSW-VS-Radio     & 10.0 & 10.7 & 11.0 & 12.2 & 12.0 &  8.7 & {\bf 22.9} \\
  GSW-BE-Bible     &  5.7 &  5.8 &  6.2 &  6.1 &  {\bf 6.4} & 6.3 & 6.3 \\
  \hline
  \end{tabular}
	  \caption{\label{tab:bleu_proposed} BLEU scores for several test sets and normalization strategies (orthographic, phonetic, character-based NMT).}
\end{table*}

Baseline1 corresponds to the system with a language model
trained only on the parallel GSW-DE data, while Baseline2 is
using a larger language model, described in
Sec.~\ref{sec:increase_data}.  We can make the following
observations:
\begin{compactitem}
\item In all the cases except GSW-BE-Novel, the orthographic approach
  improves the BLEU score of the baseline system, and the improvement
  is bigger for more remote dialects and domains.
\item The phonetic approach improves the score in 4 out of 6 cases.  In the remaining cases, we suppose that some words
  did not require pre-processing, and that the pre-processing may have
  converted the word to a false positive (i.e.\ the algorithm found a
  matching word, but it was not the correct one for translation).
\item Combining both approaches always results in better scores than
  the baseline, but in the case for which the phonetic approach score
  deteriorated, orthographic conversion only performs better.
\item In all the cases, combining CBNMT with the baseline PBSMT works
  the best.  The highest improvement is brought when dialect or domain are different (except for the Bible), because more data was used to train the CBNMT models.  This is especially true for the GSW-Archimob test set, which   has similar data as the one used to train the CBNMT models.
\item Baseline1 performs better than all the systems for
  GSW-BE-Novel test set. This is expected as the training data is both from
  the same dialect and the same domain.  Additionally, the language
  model is trained on this same data.
\end{compactitem}


\section{Conclusion}

In this paper, we proposed solutions for the machine translation of a family of dialects, Swiss German, for which parallel corpora are scarce.  Our efforts on resource collection and MT design have yielded:
  
\begin{compactitem}
\item a small Swiss German~/ High German parallel corpus of about 60k words; \item a larger list of resources which await digitization and alignment;
\item three solutions for input normalization, to address variability of region and spelling;
\item a baseline GSW-to-DE MT system reaching 36 BLEU points.
\end{compactitem}

Among the three normalization strategies, we found that character-based neural MT was the most promising one.  Moreover, we found that MT quality depended more strongly on the regional rather than topical similarity of test vs.\ training data.  

These findings will be helpful to design MT systems for spoken dialects without standardized spellings, such as numerous regional languages across Africa or Asia, which are natural means of communication in social media.

\section{Acknowledgments}

We are grateful to Swisscom for the grant supporting the first author from January to June 2017.

\section{Bibliographical References}
\label{main:ref}

\bibliographystyle{lrec}
\bibliography{mt-refs}

\end{document}